**Title:** Thickness Mapping of Eleven Retinal Layers in Normal Eyes Using Spectral Domain Optical Coherence Tomography

**Authors:** Raheleh Kafieh[1], Hossein Rabbani[1,2], Fedra Hajizadeh[3], Michael D. Abramoff [4,5], Milan Sonka[2,6]

**Meeting Presentation:** None

**Financial Support:** This work was supported in part by the National Institutes of Health grants R01 EY018853, R01 EY019112, and R01 EB004640. The sponsor or funding organization had no role in the design or conduct of this research.

**Conflict of Interest:** None of the authors have a proprietary interest.

**Abbreviated title:** SD-OCT for 11-Layer Retinal Thickness Mapping

**Address for Reprints:** Dr. Hossein Rabbani, The Iowa Institute for Biomedical Imaging, The University of Iowa, Iowa City, IA 52242, USA**, email: hossein-rabbani@uiowa.edu**

[1] Biomedical Engineering Dept., Medical Image & Signal Processing Research Center, Isfahan University of Medical Sciences, Isfahan, Iran

[2] The Iowa Institute for Biomedical Imaging, The University of Iowa, Iowa City, IA 52242, USA

[3] Vitreo-Retinal Consultant and Surgeon at Noor Eye Hospital, Noor Ophthalmology Research Center, Tehran, Iran

[4] Department of Ophthalmology & Visual Science, The University of Iowa, Iowa City, IA 52242, USA

[5] VA Medical Center, Iowa City, IA 52246, USA

[6] Department of Electrical & Computer Engineering, The University of Iowa, Iowa City, IA 52242, USA

**Summary Statement:** This study was conducted to determine the properties for thickness map of 11 retinal layers in fovea and 8 surrounding sectors. Retinal boundaries were determined automatically and the thickness results were compared between genders and for variety of ages.

**Abstract**

**Purpose.** This study was conducted to determine the thickness map of eleven retinal layers in normal subjects by spectral domain optical coherence tomography (SD-OCT) and evaluate their association with sex and age.

**Methods.** Mean regional retinal thickness of 11 retinal layers were obtained by automatic three-dimensional diffusion-map-based method in 112 normal eyes of 76 Iranian subjects.

**Results.** The thickness map of central foveal area in layer 1, 3, and 4 displayed the minimum thickness (P<0.005 for all). Maximum thickness was observed in nasal to the fovea of layer 1 (P<0.001) and in a circular pattern in the parafoveal retinal area of layers 2, 3 and 4 and in central foveal area of layer 6 (P<0.001). Temporal and inferior quadrants of the total retinal thickness and most of other quadrants of layer 1 were significantly greater in the men than in the women. Surrounding eight sectors of total retinal thickness and a limited number of sectors in layer 1 and 4 significantly correlated with age.

**Conclusion.** SD-OCT demonstrated the three-dimensional thickness distribution of retinal layers in normal eyes. Thickness of layers varied with sex and age and in different sectors. These variables should be considered while evaluating macular thickness.

## I. INTRODUCTION

Optical Coherence Tomography (OCT) is a noninvasive imaging technique that enables in-vivo cross-sectional visualization of biological tissue at micrometer resolution [1]. Low axial resolution of firstly developed OCTs (15 micrometers) made them less useful in quantitative analysis of retinal layers; however, current modalities have an improved axial resolution up to 2 micrometers. The advancement of spectral domain OCT (SD-OCT) over Standard time domain Optical Coherence Tomography (TD-OCT) provides higher speed of imaging; consequently, less eye motion artifact makes new systems able to generate 3D imaging of retina and two-dimensional thickness maps [2]. Such developments have made OCT one of the fastest adopted technologies in ophthalmology for diagnosis and study of retinal pathologies. Combination of OCT technologies with image processing and segmentation techniques provides useful information about different internal layers of retina to diagnose diseases such as glaucoma, diabetic retinopathy, and macular degeneration - the three leading causes of blindness [3]. Retinal thickness analysis is known to be an important way to quantify pathological changes [4]. According to the selection of different OCT systems for diagnosis, the under-investigation area for thickness analysis may be different, but usually around 6mm x 6mm region of macula or the optic nerve head (ONH) is selected. Several studies have reported comparisons of total retinal thickness measurements obtained by TD and SD-OCT instruments [5-7]. Many researchers focused on segmentation of retina in OCT images to produce the retinal thickness maps and to find a correlation between the quantitative and morphological features of the map, and different retinopathies [12-20].

Furthermore, thickness of Retinal Nerve Fiber Layer (RNFL) was of interest for diseases like glaucoma which is expected to change the structure of nerves in retina [21-24]. There are also some papers on extraction of thickness map for other internal layers of retina [25-34].

There is no doubt that it is useful to define a normal standard for thickness profiles which can be helpful for physicians to compare the thickness profiles of each patient with such normal sets, and also evaluate progression of certain disease that mostly involve certain retinal layers. The normative database for thickness of retina has been established in the macula [6, 7, 14, 35] and ONH regions [36]). The normative database for thickness of 3 intra-retinal layers [25], 6 intra-retinal layers [33] and choroidal thickness [32] were also reported.

In this study, we applied our previously reported 3D intra-retinal layer segmentation algorithm (using coarse grained diffusion map) [37] on SD-OCTs and normal thickness maps of 11 intra-retinal layers were generated. Similar papers like Loduca et al [33] only focus on thickness maps of 6 or less retinal layers, but our new segmentation method is able to segment 12 boundaries (11 layers) in OCT images. Furthermore, we report the correlation of age/sex of the subjects with all of the 11 layers in this research. The other important aspect of this method is its performance on 3D data, despite most of the reports [25, 33] which evaluated several 2D B-scans of OCT and combined the results to generate the thickness map. The independent standard resulted from averaging tracings from two expert observers and performance assessment results showed the mean unsigned border positioning error of 7.56±2.95 micrometers and mean signed border positioning error of -4.53±2.89 micrometers. The worst segmentation result (more error) was observed in detection of surface 4 with mean unsigned border positioning error of 12.87±4.95 micrometers and the best segmentation (less error) was found in detection of surface 7 with mean unsigned border positioning error of 4.38±1.37 micrometers. According to many similar publications, there was no particular rule to define true or false states in this segmentation; therefore sensitivity and specificity was not obtained in this study. The computation time of the proposed algorithms (implemented in MATLAB™ without using the mex files -Math Works, Inc., Natick, MA [38] was 260 seconds for each 3D volume. A PC with Microsoft Windows XP x32 edition, Intel core 2 Duo CPU at 3.00GHz, 4 GB RAM was used for calculation.

## II. MATERIAL AND METHODS

This is a cross-sectional study that subjects with normal retinal status and examination were recruited for our study. The study protocol was reviewed in advance by the review board of Noor Ophthalmology Research Center. Each participant was informed of the purpose of the study, and provided a written consent to participate. The proposed method was tested on 112 macular SD-OCT images obtained from normal eyes of 67 Iranian subjects with Heidelberg OCT- Spectralis HRA imaging system (Spectralis HRA + OCT; Heidelberg Engineering, Heidelberg, Germany) in Noor Ophthalmology Hospital. The inclusion criteria were as follows: age range between 20 and 80 years, best corrected visual acuity (BCVA) ≥20/20 , refractive error between -1 and +1 diopter, and no history or evidence of  systemic(diabetes mellitus, severe or uncontrolled systemic hypertension, pregnancy, cancer, kidney transplant, autoimmune disease ) or ophthalmic diseases such as amblyopia, high intraocular pressure (IOP; more than 21 mmHg), glaucoma and previous ocular surgery, hazy media, or poor cooperation, which prevents high-quality image acquisition. All patients underwent thorough ophthalmic examinations; including refraction, visual acuity, slit lamp biomicroscopic examination, IOP measurement by Goldman applanation tonomete, and examination of the fundus with plus 90-D lens. Nineteen sections, each composed of an average of 100 scans, were obtained within a 25×30˚ rectangle that was centered on the fovea. The resultant images were viewed and measured with the Heidelberg Eye

Explorer Software (Version 5.3). All images captured had a signal quality of at least 20 dB. Image data were saved as E2E format and transferred to Matlab Program for further analysis. The size of the obtained volumes is different from 512 ×496 × 19 to 512×496×120 voxels for directions of x, y, z (Figure 1), respectively.

In this study, we applied our previously reported 3D intra-retinal layer segmentation algorithm (using coarse grained diffusion map) [37] on SD-OCTs. This method is a fast segmentation method based on a new kind of spectral graph theory named diffusion maps. In contrast to recent methods of graph based OCT image segmentation, the presented approach does not require edge-based image information and rather relies on regional image texture. Consequently, the method demonstrates robustness in situations of low image contrast or poor layer-to-layer image gradients. Diffusion map was applied on 3D OCT datasets and for each dataset, the procedure was composed of two steps of applying diffusion maps (one for partitioning the data to important and less important sections, and another one for localization of internal layers). Diffusion maps [39] form a spectral embedding of a set $X$ of $n$ nodes, for which local geometries are defined by a kernel $k: X \times X \to R$. The kernel $k$ must satisfy $k(x,y) \geq 0$, and $k(x,y) = k(y,x)$. One may consider the kernel as an affinity between nodes which results in a graph (an edge between $x$ and $y$ carries the weight $k(x,y)$). The graph can also be defined as a reversible Markov chain by normalized graph Laplacian construction. We define

$$s(x) = \sum_y k(x,y) \qquad (1)$$

and

$$p_1(x,y) = \frac{k(x,y)}{s(x)} \qquad (2)$$

This new kernel is not symmetric, but it satisfies the requirements of being the probability of the transition from node $x$ to node $y$ in one time step, or a transition kernel of a Markov chain:

$$\forall x, \sum_y p_1(x, y) = 1 \tag{3}$$

$P$ is the Markov matrix whose elements are $p_1(x, y)$ and the elements of its powers $P^\tau$ are the probability of the transition from node $x$ to node $y$ in $\tau$ time steps. The geometry defined by $P$ can be mapped to Euclidean geometry according to eigen-value decomposition of $P$.

Such a decomposition results in a sequence of eigen-values $\lambda_1, \lambda_2, \ldots$ and corresponding eigen-functions $\psi_1, \psi_2, \ldots$ that fulfill $P\psi_i = \lambda_i \psi_i$. The diffusion map after $\tau$ time steps $\Psi_\tau: X \to R^\omega$ embeds each node $i = 1, \ldots, n$ in the Markov chain into an $\omega$-dimensional Euclidean space. Any kind of clustering like k-means may be done in this new space.

$$i \to \Psi_\tau(i) = \begin{pmatrix} \lambda_1^\tau \psi_1(i) \\ \lambda_2^\tau \psi_2(i) \\ \vdots \\ \lambda_\omega^\tau \psi_\omega(i) \end{pmatrix} \tag{4}$$

In the first step, the data pixels/voxels were grouped in rectangular/cubic sets to form a graph node. The weights of graph were also calculated based on geometric distances of pixels/voxels and differences of their mean intensity. To form the geometric distance, each element of the matrix is calculated as the Euclidean distance $\|X_{(i)} - X_{(j)}\|_2$ (or any other distance measure like Mahalanobis, Manhattan, etc.) of points; Furthermore, to construct the feature distance, the Euclidean distance of features $\|F_{(i)} - F_{(j)}\|_2$ is used [37]. The range of each distance matrix is also calculated to show how wide is the distribution of the data set and to estimate the value of the scale factor ($\sigma$). Namely, if the distance matrix $\|F_{(i)} - F_{(j)}\|_2 \left(or \ \|X_{(i)} - X_{(j)}\|_2\right)$ range from one point to another, the scale factor $\sigma_{feature}$ (or $\sigma_{geo}$) is recommended to be 0.15 times this range. This kernel can be defined as:

$$k(i,j) = \exp\left(-\frac{\|F_{(i)}-F_{(j)}\|_2^2}{2\sigma_{feature}{}^2}\right) \cdot \begin{cases} \exp\left(-\frac{\|X_{(i)}-X_{(j)}\|_2^2}{2\sigma_{geo}{}^2}\right) & \text{if } \|X_{(i)} - X_{(j)}\|_2 < r \\ 0 & otherwise. \end{cases} \quad (5)$$

where $r$ determines the radius of the neighborhood that suppresses the weight of non-neighborhood nodes and consequently makes a sparse matrix [37].

The first diffusion map clustered the data into three parts, second of which is the area of interest and the two other sections were eliminated from the next calculations. In the second step, the remaining area went through another diffusion map algorithm and the internal layers were localized based on their similarity of texture (Figure 1). Each 3D OCT images was segmented to localize 11 layers (12 surfaces) as shown in Figure 2. We used signed and unsigned errors to report the power of this method. The independent standard resulted from averaging tracings from two expert observers and performance assessment results showed the mean unsigned border positioning error of 7.56±2.95 micrometers and mean signed border positioning error of -4.53±2.89 micrometers. According to many similar publications, there was no particular rule to define true or false states in this segmentation; therefore sensitivity and specificity was not obtained in this study. Thickness maps of the 11 retinal layers were displayed in pseudo color. Total retinal thickness was also calculated by summing thickness measurements in the 11 layers.

From the retinal layer thickness map, data were grouped in 9 macular sectors within 3 concentric circles as defined by the Early Treatment Diabetic Retinopathy Study design and baseline patient characteristics (ETDRS) [40]. Figure 3 demonstrates the mentioned sectors.

In each subject, the location of the center of fovea was first identified based on the minimum thickness on the map of RNFL. This location defines the center of the concentric circles. The central circle represented the central foveal area; the second circle was sub-divided into superior (sector 2), nasal (sector 3), inferior (sector 4) and temporal (sector 5) parafoveal retinal areas. The third circle was subdivided into superior (sector 6), nasal (sector 7), inferior (sector 8) and temporal (sector 9) perifoveal retinal areas. Note that in right and left eyes, the labels 3, 5, 7, and 9 are mirrored [17].

For each of 9 sectors, and for each of 11 layers mean and SD of thickness was calculated with taking the average and standard deviation from all 112 three-dimensional datasets. Total retinal thickness was also calculated for each sector, by summing the thickness of all 11 layers.

### III. EXPERIMENTAL RESULTS

The proposed method was tested on 112 macular SD-OCT images obtained from normal eyes of 67 Iranian subjects (27 men [40%], 40 women [60%]) with Heidelberg OCT- Spectralis HRA imaging system (Spectralis HRA + OCT; Heidelberg Engineering, Heidelberg, Germany) in Noor Ophthalmology Hospital. The age of the enrolled subjects ranged from 18 to 89 years (median, 54 years and mean age, 52.9 ± 18.3). Using automatic 3D segmentation on each dataset, thickness maps of 11 equivalent of histological retinal layers in OCT were generated:

- Layer 1: which is equivalent to Nerve Fiber Layer (NFL) and containing ganglion cell axons.
- Layer 2: which is equivalent to Ganglion Cell Layer (GCL) and containing ganglion cell bodies.

- Layer 3: which is equivalent to Inner Plexiform Layer (IPL) with synaptic connections between bipolar cell axons and ganglion cell dendrites.
- Layer 4: which is equivalent to Inner Nuclear Layer (INL) with cell bodies of bipolar cells, horizontal cells, amacrine cells, interplexiform neurons, Muller cells, and some displaced ganglion cells.
- Layer 5: which is equivalent to Outer Plexiform Layer (OPL) containing synapses between photoreceptor cells and cells from the INL.
- Layer 6: which is equivalent to Outer Nuclear Layer (ONL) including rod and cone cell bodies + External (or outer) limiting membrane ELM (OR OLM) with intercellular junctions between photoreceptor cells and between photoreceptor and Muller cells and also Henle's layer (HFL) (axons of the photoreceptor nuclei).
- Layer 7: which is equivalent to Inner Segment Layer (ISL) containing cytoplasmic organelles involved in protein synthesis.
- Layer 8: which is equivalent to connecting Cilia (CL) separating the outer segment from the inner segment.
- Layer 9: which is equivalent to Outer Segment Layer (OSL) containing visual pigment in stacks of membrane discs.
- Layer 10: which is equivalent to Verhoeff Membrane (VM) including interface between cone photoreceptors and the RPE filled with photoreceptor cone outer segment tips.
- Layer 11: which is equivalent to Retinal Pigment Epithelium (RPE) which is a single layer of pigmented hexagonal cells /Bruch Membrane (BM) the innermost layer of the choroid, also called the vitreous lamina.

Examples of thickness maps generated from the left eye of one subject are shown in Figure 4. Figure 5 demonstrates the mean and SD of macular thickness in each sector for 11 retinal layers for a left eye. The location of each layer is similar to the ones in Figure 4 for simple comparison. Example of thickness map of total retina in a left eye of one subject in the study is also presented in Figure 6.

As describes above, figures 4 and 5 are depicting the retinal maps for a sample case, but the mean and SD of each layer should be calculated for the whole investigated population. For this purpose, mean and standard deviation of macular sector thickness in most important sectors and for all of 11 layers are drawn on a diagram shown in Figure 7. Intra-sector thickness variability had its largest value in layer 2 (ganglion cell layer), layer 6 (ONL+ OLM+HFL), and layer 1 (nerve fiber layer) in a descending order. In layer 1 (nerve fiber layer), the thickness variability was maximum in the perifoveal nasal area; it reduced in parafoveal (nasal and temporal) and perifoveal temporal area and the least value was seen in foveal parts. Foveal area had the highest SD in layer 2 (ganglion cell layer); the perifoveal nasal area, temporal (perifoveal and parafoveal) areas and parafoveal nasal area of second layer had the lower variations in a descending order.

The third layer (inner plexiform layer) showed a different characteristic and against all of other layers, the larger SD corresponded to perifoveal temporal segment. The variations were relatively low in this layer and the least variation occurred in foveal and para-temporal nasal area. The highest SD in thickness of the fourth layer (inner nuclear layer) was located in foveal area and the variations were lowest among the 6 first layers. The intra-sector thickness variation in $5^{th}$ layer (outer plexiform layer) had a very same pattern as the second layer (ganglion cell layer). Namely the retinal segments could be sorted according to their thickness variations as:

Foveal area, perifoveal nasal area, temporal (perifoveal and parafoveal) areas and parafoveal nasal area. Layer 6 (ONL+ OLM+HFL) had a relatively large variation in thickness, with its highest value in foveal area and the least SD value in parafoveal (nasal and temporal) segments. Intra-sector thickness variability's in layers 7 to 10 were obviously lower than first 6 layers. It is also evident that these layers were also small in their mean values. The foveal segment of the 7th layer (ISL) had the largest SD and temporal (perifoveal and parafoveal) areas, perifoveal nasal area and parafoveal nasal area had lower variations. The highest SD of the 8th layer (CL) was available in parafoveal nasal area and the other segments had similar low variations. For layers 9 (OSL) and 10 (VM), the variations were extremely low, with the highest variation in perifoveal nasal segment. Layer 11 (RPE/BM) had the largest variation among last layers. The largest variations of this layer was observed in perifoveal (nasal and temporal) areas and the standard deviation went down in foveal and parafoveal temporal regions to get the least SD in parafoveal nasal area. The numerical value of the mentioned thickness variations for each layer and in each macular sector averaged on the whole investigated populations are shown in Table 1.

Statistical analysis in this paper is based on the following rules: The effect of age on thickness of retinal layers was evaluated by hypothesis test for slope parameter of a linear regression. Variables between sexes were compared using an unpaired *t*-test. For comparing retinal thickness variables among sectors of each of 11 layers, we have multi-sample test of the means. Namely, we want to determine if the results are due to real differences, and not just sampling errors. Comparing the mean and std values obtained for each sector in each layer, we may find that mean values in a particular sector has the minimum/maximum value; however we should examine if such a minimum/maximum has statistical significance or not. For this purpose, we test between the two hypotheses:

$H_0: \mu_1 = \mu_2 = \mu_3 = \mu_4 = \cdots$

$H_1:$ At least one of them is different from the others.

We used Analysis of Variance or ANOVA (created by Fisher) as a versatile technique for dealing with this complex experimental design. In this problem we don't know what the true means are; we only have estimates from samples. So doing an ANOVA comes down to testing whether there is more variance among the samples' means than we would expect by chance sampling error alone. It results in a comparison of two variances: the variance we observed among the means of the samples versus the variance we expected to see due to sampling error. If we have these two, we can do an F test to compare these two variances statistically. We implemented the ANOVA followed by F test in Matlab program.

The total retinal thickness of the temporal and inferior quadrants was significantly greater in the men than in the women ($P<0.001$ and $P<0.002$, unpaired *t*-test). Thickness map was significantly greater in the men than in the women in many different quadrants of layer 1: the central foveal area ($P<0.001$, unpaired *t*-test) and in the inferior and temporal parafoveal area ($P<0.001$, $P<0.002$, unpaired *t*-test) and in the temporal perifoveal area ($P<0.002$, unpaired *t*-test). The thickness of other remaining retinal layers had no correlation with gender ($P>0.1$, unpaired *t*-test).

We used representative regression plots of thickness maps of retinal layers versus age to find the correlation between age and thickness of the layers. Total retinal thickness in central area did not correlate with age but in all of the remaining eight sectors it had a negative correlation with age ($P < 0.005$ for all 8 quarters, slope parameter test). Thickness of layer 1 in nasal perifoveal sector significantly correlated with age ($P<0.001$, slope parameter test); other sectors of the layer 1 tended to decay in an age-dependent manner, but not significantly ($P>0.05$, slope parameter test). Thickness of layer 4 in nasal perifoveal and nasal parafoveal sectors significantly correlated with age ($P<0.001$, $p<0.001$, slope parameter test); other sectors of the layer 4 tended to decay in an age-dependent manner, but not significantly ($P>0.05$, slope parameter test). Thickness of other remaining sectors in different retinal layers had no correlation with age ($P>0.1$, slope parameter test).

## IV. CONCLUSION AND DISCUSSION

In OCT retinal layer are recognized by their reflectivity. Although exact relationship between reflectivity and histological retinal layer is not fully understood, but quantified OCT signals have a predictable relationship to the histology and pathology in retina. Therefore, retinal thickness mapping of these layers will be useful in diagnosis and understanding of retinal pathologies. The first step in this approach is providing normative thickness maps and a complete database showing the mean value of thicknesses in each macular sector and possible variations from the mean value to determine the limitation for a person to be considered normal. Our software is designed to work properly on OCTs obtained from normal eyes. It can also produce acceptable results in case of glaucoma which doesn't alter the overall shape of OCTs [37]. However, extension of this software to analysis of OCTs from severely damaged eyes is already under investigation. In this research, we reported the thickness mean and variations for 11 layers and in each macular sector we averaged on the whole investigated normal population. The retinal layer segmentation was automatically performed using our new segmentation method [37] and most of the anatomically different layers were partitioned which sounds quite promising in diagnosis of retinal pathologies analogous to each anatomical layer.

The thickness map of layer 1 (nerve fiber layer) displayed minimum thickness in the central foveal area (P<0.001, ANOVA test) and maximum thickness nasal to the fovea (P<0.001, ANOVA test), very same as what was expected from the retinal ganglion cell axons. Retinal ganglion cells receive input from bipolar cells and amacrine cells and project their axon toward the vitreous; then the axon makes an approximately 90 degrees turn and projects toward the optic nerve head in the nerve fiber layer (Fig. 8).

The fibers in the temporal part of the retina (corresponding to the nasal visual field) course away from the fovea, and then once in the nasal retina the fibers turn back toward the optic disc, entering in the superior and inferior portions of the disc. The retinal ganglion cell axons arising from retinal ganglion cells in the nasal retina project more directly to the disc, as follows: The fibers from the nasal half of the macula, forming the papillomacular bundle (or more properly, the maculopapillary bundle), enter at the temporal disc, whereas the fibers arising from ganglion cells nasal to the disc enter at the nasal part of the disc [41]. Figure 9 shows an overall agreement between the densities of ganglion axons and the calculated RNFL thickness.

The thickness map of layer 2 (ganglion cell layer) displayed the maximum in a circular pattern in the parafoveal retinal area. This area contains ganglion cell bodies which are anatomically thicker than ganglion axons located in Layer 1. The thickness maps of layers 3 also displayed a minimum thickness in the central foveal area (P<0.001, ANOVA test). This occurs due to natural tendency of cells backward the fovea to permit the light pass toward the photoreceptor cells. The thickness map of layer 3 has also the maximum thickness in a circular area in the parafoveal retinal area, but the maximum value and the regularity of thickness in circular region are both less that layer 2. Layer 3 includes synaptic connections between bipolar cell axons and ganglion cell dendrites, the structure of low thickness in the fovea could be expected since the dendrites of cells are expected to have a similar arrange with axon (layer 1) of the same cells (ganglion cells); besides, the dendrites has a predictably less thickness in comparison with cell bodies (layer 2). The thickness map of layer 4 (inner nuclear layer) showed a minimum thickness in the central foveal area (P<0.005, ANOVA test) and the maximum thickness was seen in a circular area in the parafoveal retinal area. Similar to layer 3, the maximum value and the regularity of thickness in circular region are both less that layer 2. Layer 4 contains cell bodies of bipolar cells, horizontal cells, amacrine cells, interplexiform neurons, and Muller cells; therefore, it could be expected to show a higher thickness than synapses located in Layer 5. Thickness of the layer 5 (outer plexiform layer) was low particularly in the perifoveal nasal area (P<0.001, ANOVA test). Containing synapses between photoreceptor cells and cells from the inner nuclear layer, thickness of this layer could be anticipated lower than previously discussed cell bodies. Thickness of layer 6 (ONL+ OLM+HFL) was highest in the central foveal area (P<0.001, ANOVA test). It is well known that the fovea contains the highest density of cone photoreceptors in the retina; the numbers drop to about 50% by 500 micrometers from the fovea centre and to less than 5% at about 4 mm eccentricity [42]. Cone density may increase slightly in the far nasal retina, however [43]. Therefore Layer 6 including rod and cone cell bodies shows admissibly high thickness in comparison to other layers (P<0.001, ANOVA test).

The thickness map of layer 7 (ISL) displayed a uniform and low thickness in all retinal areas. The thickness map of layer 8 (CL), layer 9 (OSL), and layer 10 (VM) showed a relatively uniform with low thickness. Photoreceptor inner segments (ISL, layer 7) contain the support organelles (mitochondria, ribosomes, endoplasmic reticulum, synaptic vesicles, etc), and the axon terminal (where neurotransmitter is released). The capture of individual photons by the photopigment molecules in the disk membranes is what initiates neural signaling. Photoreceptors are actually specialized hair cells. The inner and outer segments are connected by the cilium layer (CL, Layer 8). The outer segment (OSL, Layer 9) contains photo-pigment in free-floating disks (rods) or folded layers (cones). Cone outer segments have a continuous outer membrane, whereas rods have discs, stacked like coins, in a sleeve. The rod and cone outer segment membranes are constantly being replenished (like fingernails, they just keep growing). This is why the pigment epithelium must trim off the excess, a process known as phagocytosis. Finally, VM (Layer 10) includes the interface between cone photoreceptors and the RPE. The above mentioned 4 layers have similar characteristics that exhibit smooth and flat characteristics in normal eyes; however, as discussed in the rest of paper, their smooth presentation may be exposed to change in different pathological situations. The thickness map of layer 11 (RPE/BM) had a semi-constant pattern and the mean thickness was relatively low but higher than layers 7 – 10 (P<0.005, ANOVA test). The retinal pigment epithelium (RPE) is a monolayer of pigmented cells forming a part of the blood/retina barrier [44]. The apical membrane of the RPE faces the photoreceptor outer segments. Long apical microvilli surround the light-sensitive outer segments establishing a complex of close structural interaction. With its basolateral membrane the RPE faces Bruch's membrane, which separates the RPE from fenestrated endothelium of the choriocapillaris. The nodular and coarse anatomical morphology of this layer is also in agreement with calculated thickness maps.

According to our findings, thickness maps of the nerve fiber layer (layer 1), IPL (layer 3) and INL (layer 4) had the lowest value in the foveal area. In agreement with findings of Loduca [33] the nerve fiber layer thickness was highest in the perifoveal retinal area (near the optic nerve head), due to the high density of nerve fiber bundles near the optic nerve head. Furthermore, the nerve fiber layer had very high variations in thickness after layer2 (ganglion cell layer) and layer 6 (ONL+ OLM+HFL). This study can segment the GCL and inner plexiform layer individually, which were merged in the segmentation method by Loduca [33]. The GCL (layer 2) had the highest value in parafoveal retinal area, particularly nasal to the fovea due to the existence of ganglion cell bodies. The IPL (layer3) had also the highest value in parafoveal retinal area, but particularly superior and inferior to the fovea. Loduca [33] reported that intra-sector thickness variability of GCL + IPL was largest compared to all other layers; now, we can deliberately claim that the thickness variability the GCL (layer 2) is the largest, but the IPL (layer 3) has a low variation in thickness. Confirming the reports of Loduca [33], we found that the INL(layer 4) had a minimum in the fovea area and the thickness variations were relatively low. Thickness of the OPL (layer 5) was similar to findings of Loduca [33] and showed a uniform and smooth structure with a restively low thickness variation. In this research we are also able to segment the ONL+ OLM+HFL (layer 6) from ISL (layer7) which were merged and named the ONL+ photoreceptor inner segments by Loduca [33]. For ONL+ OLM+HFL(layer 6), an obvious maximum occurred in fovea area due to existence of the highest density of cone photoreceptors. The thickness variations of layer 6 was also very high (after layer2 (ganglion cell layer)). The thickness of ISL (layer7) was found to be uniform with slight variations. In the 8th layer (CL), 9th layer (OSL) and 10th layer (VM), the thickness was uniform and the variations were low. In Layer 11 (RPE/BM), the thickness was more than other layers of photoreceptor outer segment and the variation in thickness was relatively high. To demonstrate the similarity of our findings to Loduca [33], we made the thickness maps of 6 layers proposed in [33], by combining the results of GCL (layer 2) and IPL(layer 3) to form the layer 2 of Loduca, combining the results of ONL+ OLM+HFL (layer 6) and ISL (layer7) to form the layer 5 of Loduca, and combining the results of CL (layer 8), OSL (layer 9), VM (layer 10) and RPE/BM (layer 11) to form the layer 6 of Loduca in figure

10. This figure shows the high similarity of results and confirms more details are available in the current study in comparison to [33]. Table 2 is a comparison of total retinal thickness measurements in 9 macular sectors obtained using different optical coherence tomography instruments and the current study. The last column is a Time Domain OCT and the rest are Spectral Domain OCTs. As it could be expected, the values obtained by Time Domain OCT are lower than other studies, but the results obtained from Spectral Domain OCTs are similar. The correct calculation of thickness maps for different retinal layers is promising in diagnosis of retinal pathologies. Corresponding to ganglion cell axons, bodies, and dendrites, layers 1, 2, and 3 are of real importance for studying pathologies pertaining to ganglion cells like glaucoma. The separate study of these layers (presented in this paper) can even clarify the onset of anatomical change in distinctive parts of ganglion cells. Layers 3, 4, and 5 are prominent in detection of diabetes and retinal vascular diseases. Since blood delivery in one third of internal retina (from retinal artery) and two third of external area (from choroidal zone) are different, the middle region (Layers 3, 4, and 5) can be considered as watershed and vascular diseases can be diagnosed through them. Layer 6 is undoubtedly consequential in detection of healthiness of photoreceptors and possibly in identification of visual acuity. This layer may be disrupted in hydroxychloroquine poisonings, trauma, eye inflammations, macular heredity diseases and central serous retinopathies. The ONL thickness is also positively correlated with the Best-Corrected Distance Visual Acuity (BCVA) in resolved central serous chorioretinopathy (CSC) [45].

Layers 7, 8, 9, 10 are less investigated by other researches. CL(layer 8) as a junction of IS and OS layers is reported to be important in prediction of poor visual acuity among patients with epiretinal membranes (ERM) [46] and measurement of retinal sensitivity in both dry and wet forms of AMD [47]. The VM (Layer 10) is also investigated in the subfoveal area of the high-resolution images of X-linked retinoschisis (XLRS) to be invisible, suggesting a disruption within the cone photoreceptors and the RPE [48]. The thickness of these 4 layers should also be carefully inspected in different pathologies to get useful information on their correlation. Layer 11 is the best indicator of Age related Macular Degeneration (AMD) and can also be evaluated in geographic atrophies. Furthermore, dysfunction of RPE in diseases like Retinitis Pigmentosa and loss of pigment in the eyes of albinos, are considerable applications of studying on thickness of this layer.

Table 1- Intra-sector thickness variability in 9 macular sectors and in 11 retinal layers

| | Mean(std) of Standard Deviations (µm) | | | | | | | | |
|---|---|---|---|---|---|---|---|---|---|
| | Sector 1 | Sector 2 | Sector 3 | Sector 4 | Sector 5 | Sector 6 | Sector 7 | Sector 8 | Sector 9 |
| Layer 1 | 8(2) | 10(3) | 9(2) | 11(3) | 9(2) | 16(4) | 13(4) | 12(3) | 6(2) |
| Layer 2 | 23(4) | 13(3) | 12(2) | 14(2) | 15(3) | 16(3) | 18(3) | 14(3) | 14(2) |
| Layer3 | 3(1) | 4(1) | 3(1) | 8(2) | 7(1) | 6(1) | 6(1) | 6(1) | 6(1) |
| Layer 4 | 12(2) | 4(1) | 4(1) | 5(1) | 4(1) | 8(2) | 6(1) | 6(1) | 5(1) |
| Layer 5 | 12(3) | 8(1) | 6(1) | 9(1) | 6(1) | 7(1) | 8(2) | 6(1) | 7(1) |
| Layer 6 | 17(3) | 9(1) | 9(2) | 7(1) | 7(1) | 10(2) | 15(3) | 10(1) | 10(2) |
| Layer7 | 7(1) | 3(1) | 4(1) | 4(1) | 5(1) | 5(1) | 5(1) | 4(1) | 4(1) |
| Layer 8 | 5(1) | 4(1) | 5(1) | 2(0) | 1(0) | 4(1) | 4(1) | 2(0) | 3(1) |
| Layer 9 | 3(1) | 2(1) | 1(0) | 3(1) | 1(0) | 3(1) | 4(1) | 2(0) | 3(1) |
| Layer 10 | 3(1) | 2(0) | 1(0) | 2(0) | 1(0) | 4(1) | 2(0) | 3(1) | 3(1) |
| Layer 11 | 5(1) | 4(1) | 4(1) | 3(1) | 5(1) | 6(1) | 7(1) | 5(1) | 7(1) |

The 11 retinal layers were the NFL (layer 1), GCL (layer 2), IPL (layer 3), INL (layer 4), OPL (layer 5), ONL + OLM + HFL (layer 6), ISL (layer7), CL (layer 8), OSL (layer 9), VM (layer 10), RPE/BM (layer 11). Sector 1 = fovea; Sector 2 = parafoveal superior, Sector 3 = parafoveal nasal, Sector 4 = parafoveal nferior, Sector 5 = parafoveal temporal; Sector 6 = perifo-veal superior, Sector 7 = perifoveal nasal, Sector 8 =perifoveal inferior, Sector 9 = perifoveal temporal.

Table 2- comparison of total retinal thickness measurements in 9 macular sectors obtained using different optical coherence tomography instruments (mean (SD) of thickness measurements)

| | Spectralis, HRA, Heidelberg Engineering, Germany [†] | Spectralis, Heidelberg Engineering, Germany [‡] | Cirrus HD-OCT, Carl Zeiss Meditec, Inc [**] | RTVue-100, Optovue, Inc. fremont, CA., [**] | 3D OCT-1000, Topcon, Inc., Paramus, NJ [**] | 3D OCT-1000, Topcon, Inc. [††] | Stratus OCT [OCT3], Carl Zeiss Meditec, Dublin, Calif [‡‡] |
|---|---|---|---|---|---|---|---|
| Macular sector | | | | | | | |
| Fovea | 265(15) | 264 (13) | 262 (16) | 256 (15) | 231 (16) | 222 (18) | 212 (20) |
| **Parafovea** | | | | | | | |
| Temporal | 307(15) | 306 (14) | 306 (10) | 308 (13) | 280 (10) | 285 (14) | 251(13) |
| Superior | 312(13) | 311 (14) | 320 (12) | 324 (11) | 293 (12) | 297 (15) | 255(17) |
| Nasal | 325(16) | 325 (15) | 323 (12) | 324 (11) | 296 (12) | 299 (15) | 267(16) |
| Inferior | 316(14) | 314 (13) | 316 (11) | 318 (10) | 288 (10) | 294 (15) | 260(15) |
| **Perifovea** | | | | | | | |
| Temporal | 255(11) | 256 (14) | 255 (9) | 265 (10) | 234 (16) | 244 (12) | 210(14) |
| Superior | 273(17) | 271 (18) | 274 (13) | 278 (13) | 249 (13) | 257 (13) | 239(16) |
| Nasal | 272(17) | 298 (19) | 293 (13) | 291 (14) | 266 (13) | 273 (14) | 246(14) |
| Inferior | 265(16) | 274 (17) | 264 (11) | 267 (12) | 240 (12) | 247 (13) | 210(13) |

[*]Mean (SD) of thickness measurements are shown in μm.
[†]Current study.
[‡] Loduca and associates[37]
[**]Sull and associates[6]
[††]Ooto and associates[39]
[‡‡]Chan and associates[17]

**Figures:**

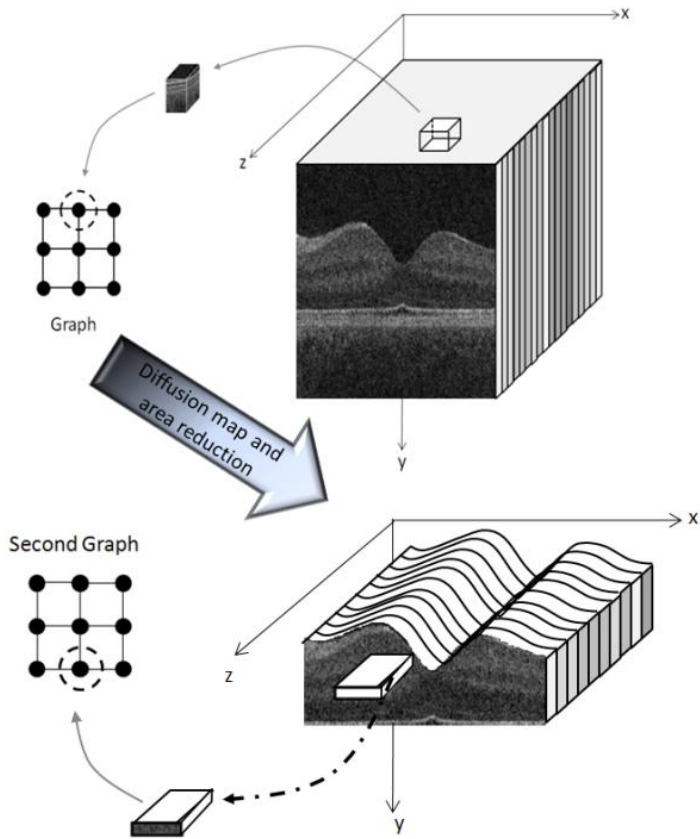

Fig. 1. Construction of graph nodes from a 3D OCT.

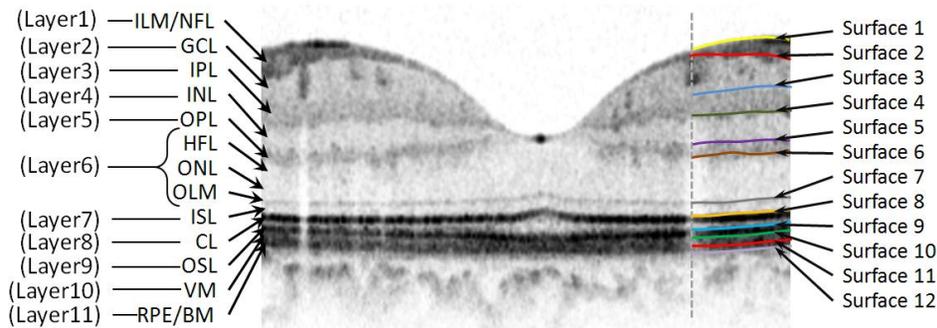

Fig. 2. Segmentation on the original image. Retinal layers are: Inner Limiting Membrane (ILM), Nerve Fiber Layer (NFL), Ganglion Cell Layer (GCL), Inner Plexiform Layer (IPL), Inner Nuclear Layer (INL), Outer Plexiform Layer (OPL), Henle's layer (HFL), Outer Nuclear

Layer (ONL), Outer limiting membrane (OLM), Inner Segment Layer (ISL), connecting Cilia (CL), Outer Segment Layer (OSL), Verhoeff Membrane (VM), Retinal Pigment Epithelium (RPE), Bruch Membrane (BM).

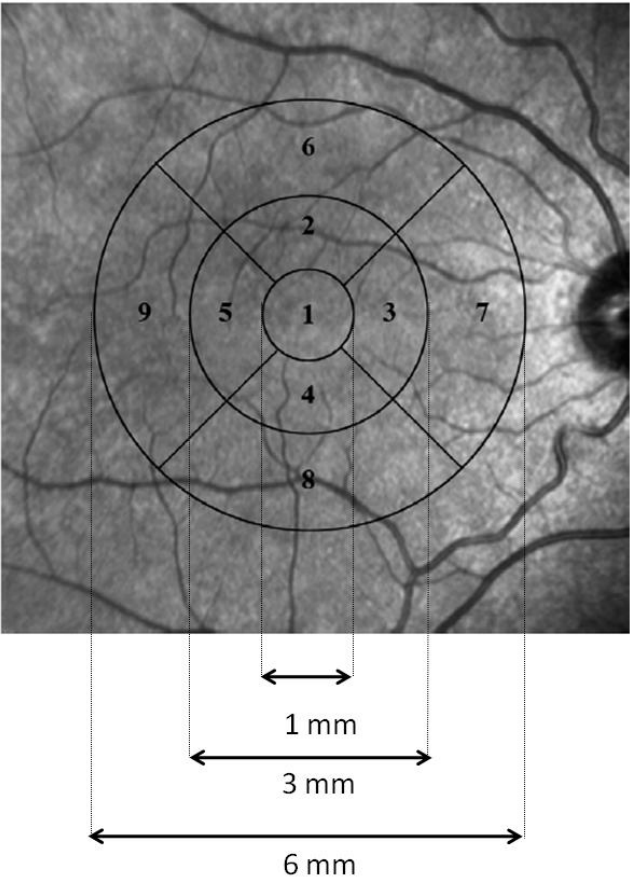

Fig. 3. ETDRS grid used for reporting retinal thickness [40].

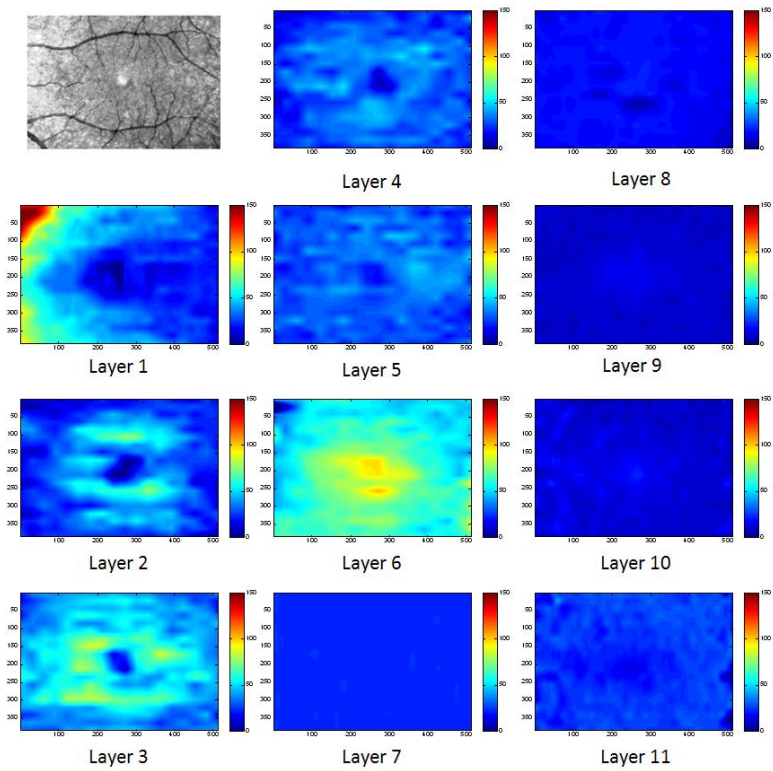

Fig. 4. Examples of thickness maps of 11 retinal layers in a right eye of one subject in the study. The 11 retinal layers were the NFL (layer 1), GCL (layer 2), IPL (layer 3), INL (layer 4), OPL (layer 5), ONL + OLM+HFL (layer 6), ISL (layer7), CL (layer 8), OSL (layer 9), VM (layer 10), RPE/BM (layer 11).

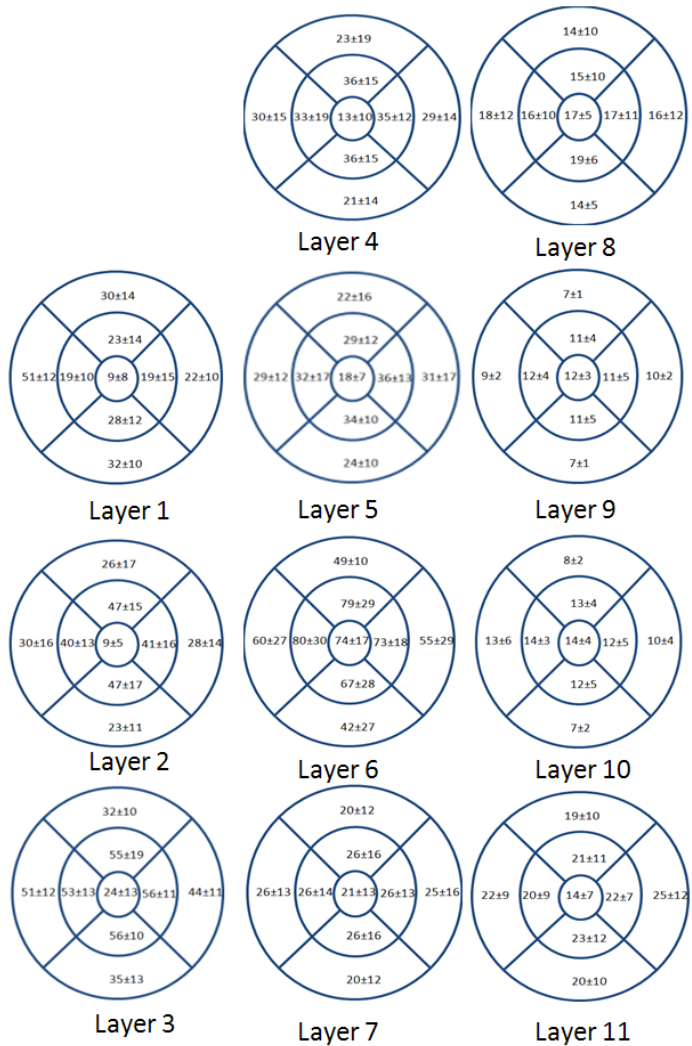

Fig. 5. The mean and SD of macular thickness in each sector for 11 retinal layers (for a right eye). The 11 retinal layers were the NFL (layer 1), GCL (layer 2), IPL (layer 3), INL (layer 4), OPL (layer 5), ONL + OLM+HFL (layer 6), ISL (layer 7), CL (layer 8), OSL (layer 9), VM (layer 10), RPE/BM (layer 11).

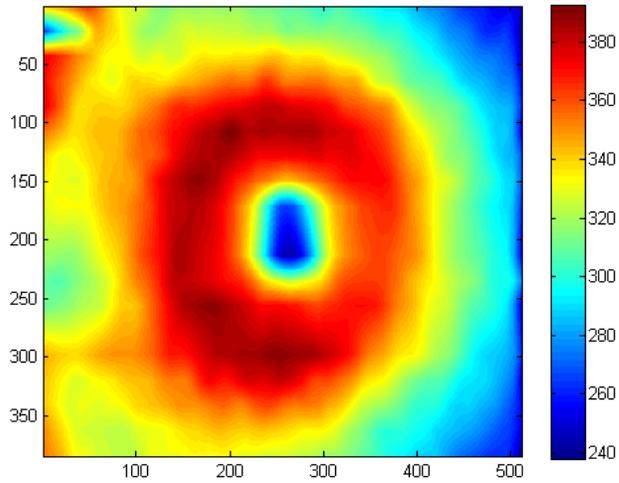

Fig. 6. Example of thickness map of total retina in a right eye of one subject in the study.

Fig. 7. Mean and standard deviation of macular sector thickness plotted as a function of retinal eccentricity for 11 retinal layers. The 11 retinal layers were the NFL (layer 1), GCL (layer 2), IPL (layer 3), INL (layer 4), OPL (layer 5), ONL + OLM+HFL (layer 6), ISL (layer7), CL (layer 8), OSL (layer 9), VM (layer 10), RPE/BM (layer 11).

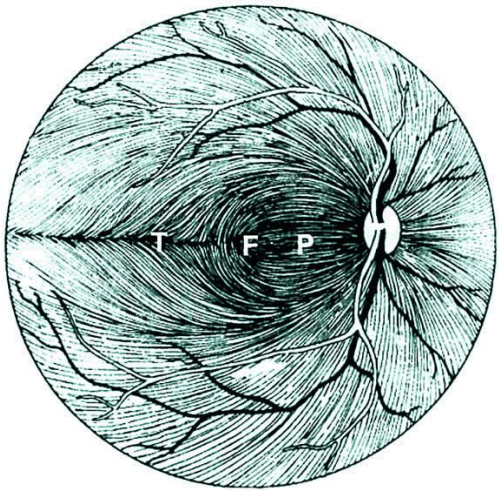

Fig. 8. Course of the retinal ganglion cell axons within the nerve fiber layer of the retina. (F, fovea; P, papillomacular bundle; T, temporal raphe) [49].

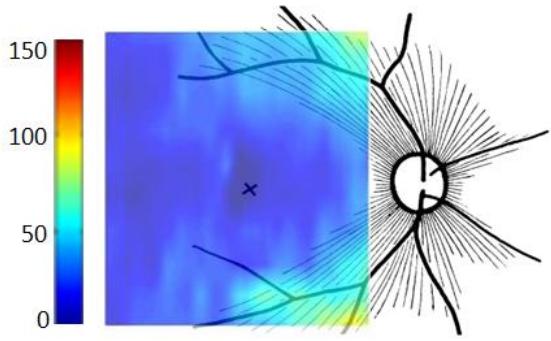

Fig. 9. Agreement between schematic representation of retinal radial peripapillary capillaries in man (X=fovea) (Reproduced from [50]) and the calculated RNFL thickness. The fibers in the temporal part of the retina (corresponding to the nasal visual field) course away from the fovea, and then once in the nasal retina the fibers turn back toward the optic disc, entering in the superior and inferior portions of the disc. The fibers crowded with higher thickness in superior and inferior as they become closer to optic disk.

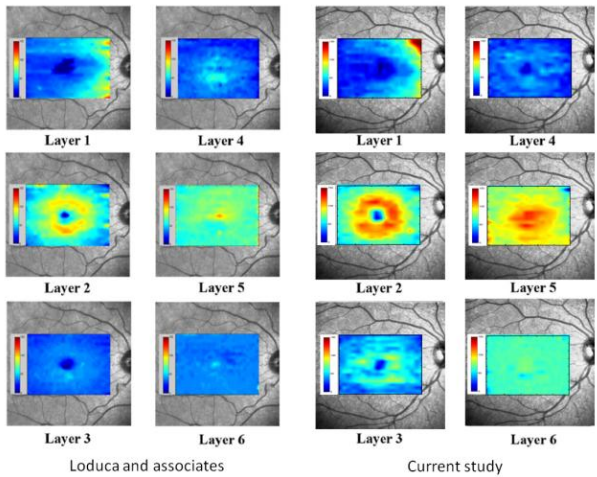

Fig. 10. Comparison of the thickness maps of 6 layers proposed in [33] to the thickness maps of those 6 layers in our study by combining the results of GCL (layer 2) and IPL(layer 3) to form the layer 2 of Loduca, combining the results of ONL+ OLM +HFL(layer 6) and ISL (layer7) to form the layer 5 of Loduca, and combining the results of CL (layer 8), OSL (layer 9), VM (layer 10) and RPE/BM (layer 11) to form the layer 6 of Loduca